\documentclass[10pt,twocolumn,letterpaper]{article}
\pdfoutput=1

\usepackage{cvpr}              

\usepackage{graphicx}
\usepackage{amsmath}
\usepackage{amssymb}
\usepackage{booktabs}
\usepackage{algorithmic}
\usepackage{multirow}
\usepackage{booktabs}
\usepackage{xr}
\usepackage{xcolor}

\usepackage[pagebackref,breaklinks,colorlinks]{hyperref}

\usepackage[capitalize]{cleveref}
\crefname{section}{Sec.}{Secs.}
\Crefname{section}{Section}{Sections}
\Crefname{table}{Table}{Tables}
\crefname{table}{Tab.}{Tabs.}

\def\cvprPaperID{8233} 
\def\confName{CVPR}
\def\confYear{2022}

\begin{document}

\title{NightLab: A Dual-level Architecture with Hardness Detection for\\ Segmentation at Night}


\newcommand*{\affmark}[1][*]{\textsuperscript{#1}}
\newcommand*{\email}[1]{\small\texttt{#1}}
\newcommand*{\affaddr}[1]{#1}

\author{Xueqing Deng\affmark[1,2]\thanks{This work was done when Xueqing was interned at ByteDance.}, Peng Wang\affmark[2], Xiaochen Lian\affmark[2], Shawn Newsam\affmark[1]\\
\affaddr{\affmark[1]EECS, University of California at Merced, \affaddr{\affmark[2]ByteDance Inc.}}\\
\email{\{xdeng7, snewsam\}@ucmerced.edu}, \email{\{peng.wang, xiaochen.lian\}@bytedance.com}
}

\maketitle

\begin{abstract}
The semantic segmentation of nighttime scenes is a challenging problem that is key to impactful applications like self-driving cars. Yet, it has received little attention compared to its daytime counterpart. In this paper, we propose \texttt{NightLab}, a novel nighttime segmentation framework that leverages multiple deep learning models imbued with night-aware features to yield State-of-The-Art (SoTA) performance on multiple night segmentation benchmarks. Notably, \texttt{NightLab} contains models at two levels of granularity, i.e. image and regional, and each level is composed of light adaptation and segmentation modules. Given a nighttime image, the image level model provides an initial segmentation estimate while, in parallel, a hardness detection module identifies regions and their surrounding context that need further analysis. A regional level model focuses on these difficult regions to provide a significantly improved segmentation. All the models in \texttt{NightLab} are trained end-to-end using a set of proposed night-aware losses without handcrafted heuristics. Extensive experiments on the NightCity~\cite{tan2020nightcity} and BDD100K~\cite{yu2018bdd100k} datasets show \texttt{NightLab} achieves SoTA performance compared to concurrent methods. Code and dataset are available at \href{https://github.com/xdeng7/NightLab}{https://github.com/xdeng7/NightLab}.
\end{abstract}

\section{Introduction}
\label{sec:introduction}
\vspace{-3pt}
\begin{figure}
    \centering
    \includegraphics[width=\linewidth]{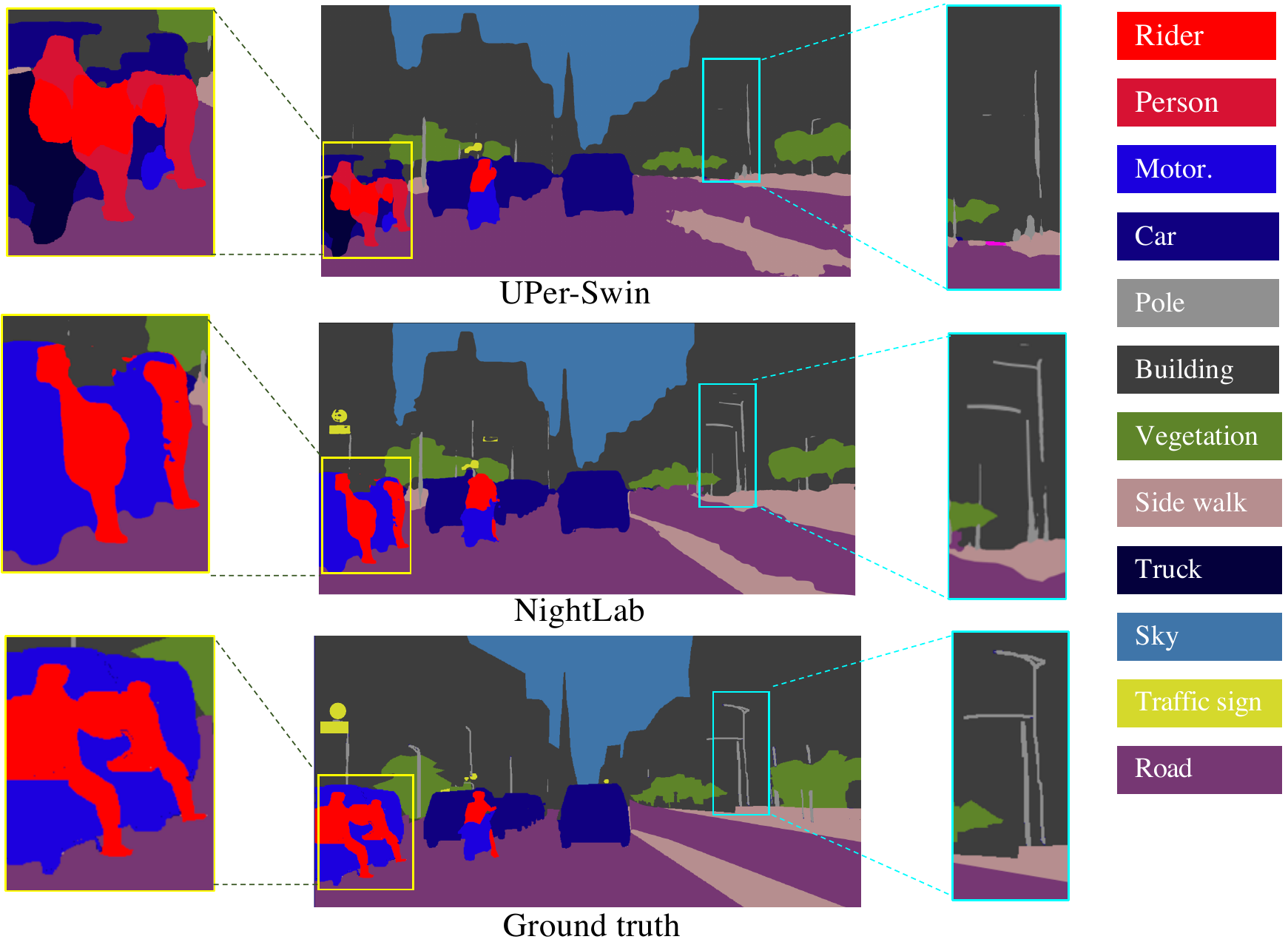}
    \vspace{-20pt}
    \caption{Visual comparisons of segmentation results from UPer-Swin\cite{liu2021swin} and our proposed \texttt{NightLab}. \texttt{NightLab} shows improvements on the parts of motorcycle and rider, where UPer-Swin predicts rider as person, and motorcycle as car, and poles are missing. \texttt{NightLab} is able to provide details for small objects. }

\vspace{-5pt}
    \label{fig:overview}
\end{figure}
\begin{figure*}[t]
    \centering
    \includegraphics[width=0.9\linewidth]{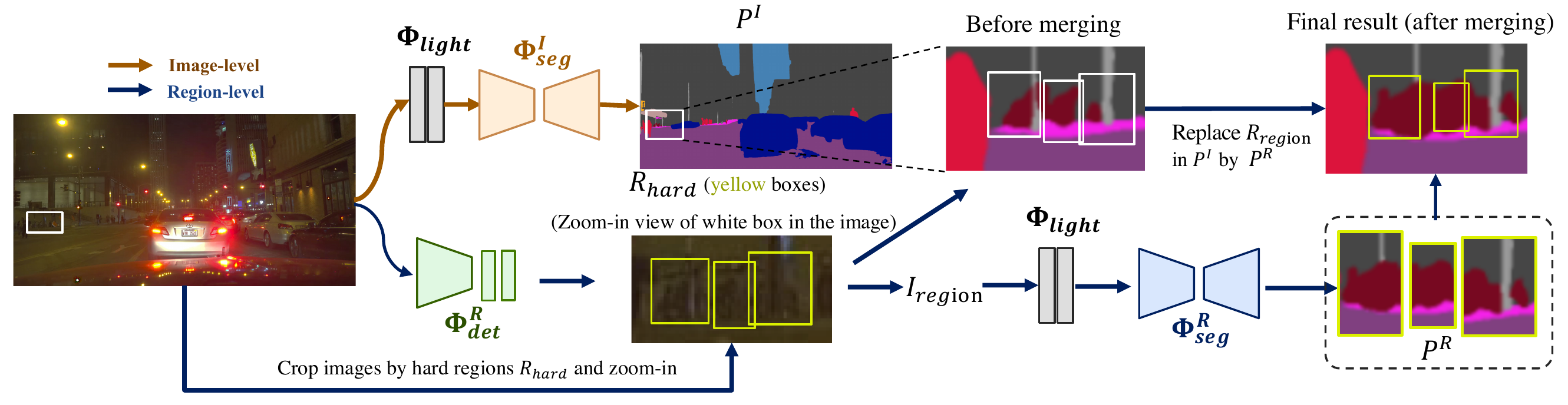}
     \vspace{-5pt}
    \caption{\textbf{\texttt{NightLab} overview and inference}. With the input images, there is a duel-level architecture to produce the final output. Note the hard contexts (boxes) here are automatically discovered without ground truth. 1) The image-level networks $\Phi_{light}^I, \Phi_{seg}^I$ is used to create predictions $P^I$ for the whole images. Most of the easy regions $\mathcal{R}_{easy}$ can be accurately predicted by $\Phi_{seg}^I$. 2) Then, hard regions $\mathcal{R}_{hard}$ will be detected by HDM $\Phi_{det}^R$ with the input images. Once the regions are discovered, they will be zoom-in and processed by $\Phi_{light}^R$ and $\Phi_{seg}^R$ to obtain local prediction $P^R$ of $\mathcal{R}_{hard}$. At last, $P^R$ will be merged back to $P^I$ to generate the final segmentation output. }
    \vspace{-5pt}
    \label{fig:deployment}
\end{figure*}
Semantic segmentation is a fundamental task in computer vision on which there has been much progress recently with the introduction of deep semantic parsing methods, \eg, DeepLab~\cite{chen2017deeplabv2,chen2018deeplabv3plus} and Transformers~\cite{dosovitskiy2020vit, liu2021swin}. However, the focus has been almost entirely limited to daytime benchmarks such as CityScapes~\cite{Cordts2016cityscapes} and ADE20k~\cite{zhou2017ade20k}. Much less progress has been made on the nighttime problem such as establishing strong benchmarks and designing effective architectures. Yet, success on the nighttime scene segmentation is crucial for a number of impactful applications such as autonomous driving~\cite{menze2015autonmous_driving}, robotic vision~\cite{desouza2002robotics}, etc. 


There are far fewer open-source labelled nighttime images than daytime ones. Most nighttime image collections contain only unlabeled images and so there has been a lot of work ~\cite{wu2021dannet} on unsupervised domain adaptation between daytime and nighttime for segmentation. Our experience, based on experiments, is that these adaptation frameworks perform poorly in practice due to the large domain gap between daytime and nighttime scenes.


Recently, Tan \etal~\cite{tan2020nightcity} proposed NightCity which makes progress on two key challenges in  nighttime segmentation: the lack of a large realistic dataset and the large illumination variation that results from over or under exposure in night scenes. The NightCity effort resulted a large dataset of densely labelled images and a segmentation model that contains an exposure-guided layer designed for light changes. The model is shown to outperform unsupervised methods.


Our work in this paper takes additional steps in this direction and proposes \texttt{NightLab}, a nighttime segmentation framework focused on architecture optimization using real labelled night images which results in a significant, \ie, $\sim$10\%, absolute improvement over the original NightCity baseline~\cite{tan2020nightcity}. Specifically, we employ effective model architecture design to achieve two goals related to the large illumination variation in nighttime images. First, is to reduce the amount of light variation. Rather than performing simple exposure enhancement, we propose a regularized light adaptation module (ReLAM) based on a large amount of day and night images. Different from image translation approaches~\cite{isola2017pix2pix,zhu2017cyclegan, Anokhin_2020_CVPR} that can significantly alter image appearance, ReLAM preserves night image texture which helps avoid large domain shifts during adaptation, yielding better generalization for night images. Second, due to the low-light levels and blurry texture, small objects are often not distinguishable based on their appearance alone. Therefore, as illustrated in early work on object understanding~\cite{oliva2007role}, context is crucial for helping resolve potential ambiguity in certain nighttime image regions. While most deep networks contain multi-scale structures by enumerating scales such as HRNet~\cite{wang2020hrnet}, night images have objects with substantial scale variation, \eg, road light and bicycle as illustrated in Fig.~\ref{fig:overview}. Such variation is often beyond the scope of the enumerated scales in modern networks. To tackle this issue, we propose a hardness detection module (HDM), which adopts the idea of regional proposal network (RPN) from objection detection. Our HDM identifies regions, along with their context, that need additional attention and analysis. Finally, we adopt the SoTA architecture of Swin-Transformer~\cite{liu2021swin} as our segmentation encoder and embed DeformConv~\cite{dai2017deformable} as the decoder. This provides improved architecture capacity and context modeling ability.

In summary, as illustrated in Fig.~\ref{fig:deployment}, inference in \texttt{NightLab} works as follows. Given a night image, the image level model first adapts the image light through ReLAM ($\Phi_{light}^I$) and sends it to an image-level segmentation model ($\Phi_{seg}^{I}$), producing an overall segmentation. In tandem, HDM is used to detect hard regions that need further analysis. These regions are cropped, batched, and sent to a regional level model which adapts and segments these regional patches similar to the image level. Here, our regional model is not trained over the full set of classes but is limited to a subset of automatically identified difficult classes such as bicycle and road light to better mine the context information needed to distinguish their semantics. The segmented results from the region level are then merged with the image level parsing results, yielding the final segmentation. 


Finally, since we found many mislabelled pixels in the NightCity validation images as shown in Sec.~\ref{sec:experiment}, we manually relabel the dataset so that the evaluation of our and other methods is more meaningful. Extensive experimentation shows \texttt{NightLab} outperforms concurrent methods and ablative studies demonstrate the contribution of each of the proposed modules


In summary, our contribution includes:

\begin{enumerate}
\vspace{-8pt}
    \item We propose \texttt{NightLab}, a dual-level architecture with novel modules including ReLAM and HDM specifically designed for night scene segmentation. The framework achieves SoTA performance on multiple nighttime benchmarks.
    \vspace{-8pt}
    \item We propose an effective training pipeline for the architecture whose modularity provides good interpretability of our improvements.
    \vspace{-8pt}
    \item We derive a more accurate benchmark dataset from NightCity and conduct extensive experiments that investigate a variety of night scene segmentation strategies. Our benchmark dataset and strong baseline serve as a good starting point for future researchers.
\end{enumerate}

\section{Related Works}
\vspace{-5pt}
\label{sec:related_works}

\noindent{\textbf{Semantic segmentation.}} This task has been actively studied in past few decades, and turns to be practical in many real-world applications with the rising of deep learning with convolutions~\cite{zhu2021pami, peng2015depth, xia2016zoom, zhu2019cvpr, zhu2021unified}. In general, two principles are followed when designing the architecture,\ie, discover multi-scale context and design high-resolution representation.  Some representative works includes the initial Fully Convolutional Network (FCN) based methods \cite{long2015fully}, the series of DeepLab networks \cite{chen2017deeplabv2,chen2017deeplabv3, chen2018deeplabv3plus}, multi-scale aware networks like HRNet~\cite{wang2020hrnet},  PSPNet~\cite{zhao2017pspnet},  and models with attention module such as cross attention~\cite{huang2019ccnet}.  To better model object context, Deformable convolution~\cite{dai2017deformable} is proposed to be embedded in these network architecture for performance enhancement. 

Most recently, Transformer~\cite{devlin2018bert} shows advanced performance due to its multi-layer full attention mechanism in language processing. DPT~\cite{Ranftl2020dpt} first shows a full transformer based network which outperforms convolutional based architectures. Later, transformers have raised the attention in the computer vision community. Vision Transformers\cite{dosovitskiy2020vit, xie2021segformer,wang2021crossformer, lee2021vitgan,touvron2021deit,carion2020detr} have been widely studied for various vision tasks.  Most recently, to reduce the computational complexity inside transformer, Swin-Transformer~\cite{liu2021swin} proposes a shifted window operation, which provides SoTA performance over various  benchmarks. 
In our work, we adopt Swin-Transformer as our backbone, and show it significantly improves over our night segmentation benchmarks from NightCity~\cite{tan2020nightcity}. 
However, there still remain issues brought by light variation in nighttime motivating us to design various modules to enhance its performance.





\noindent{\textbf{Domain adaptation (DA) for segmentation.}} DA is designed for transferring knowledge from source to target domain, where usually there are rich labeled data in source domain while unlabeled data in target domain. For example, lots of works~\cite{Tsai_adaptseg_2018, tsai2019patch,vu2019advent, luo2019taking, pan2020intrada, sankaranarayanan2018learning, zhu2018penalizing} try to adapt segmentation model trained from synthetic images, \eg, GTA5~\cite{eccv2016gta5} or SYNTHIA\cite{ros2016synthia},  to real images, \eg, Cityscapes. Instead of adopting models, some works~\cite{hoffman2018cycada,wu2018dcan, eccv2020forkgan, eccv2018auggan, cvpr2021comogan,ijcv2021dlow, lin2020multimodal} try to adapt images by applying style transfer\cite{zhu2017cyclegan} that transforms images in target domain to source domain. The former is trying to obtain a domain-invariant representations which has to be retrained when a new domain is added. The latter does not need to change the segmentation network but only need to train a new adaptor. 
Specifically, when targeting at daytime and nighttime adaptation,  Song \etal \cite{TITS2020nighttime} follows the former strategy, which proposes to transfer unlabeled day-time and night-time images into a shared latent feature space. Sun \etal \cite{sun2019daynightconvert} follows the latter by proposing to translate the day-time and night-time images by CycleGAN, and training the segmentation model on synthetic night-time images.
Similar approaches using CyclanGAN such as using a curriculum framework for adaptation\cite{sakaridis2020map, iccv19gcma} is explored across different time (daytime, twilight, and nighttime).   
However, existing domain adaptation methods for segmentation are mostly in a unsupervised manner.  The improvement of adaptation will be significantly marginalized when supervised label is available. 
In \texttt{NightLab}, our architecture falls in the strategy of adaptation then segmentation since it has better explain ability, and we carefully design a regularized module, \ie, ReLAM, to make it useful in supervised manner.


\noindent{\textbf{Vision tasks in the dark.}}
Meanwhile, there are rising interests in analyzing images in the dark, \eg, localization\cite{icra2019night_localization}, depth estimation\cite{iccv2021regularizing}, object detection\cite{tiv2020night_detection}, person reidentification\cite{iccv2019lightaug,romera2019zju}, etc. 
Instead of using synthesized images, the major approaches are working on various \textit{enhancement}, which try to lighten the low-illumination areas and scenes for easier feature extraction. For example, \cite{iccv2021regularizing} leverages mapping consistent image enhancement module to enhance image visibility for depth estimation. \texttt{NightLab} set up a new benchmark in the field of segmentation, and we hope our approach could benefit mutually with other tasks such as depth or video understanding.


\begin{figure*}[!htp]
    \centering
    \includegraphics[width=\linewidth]{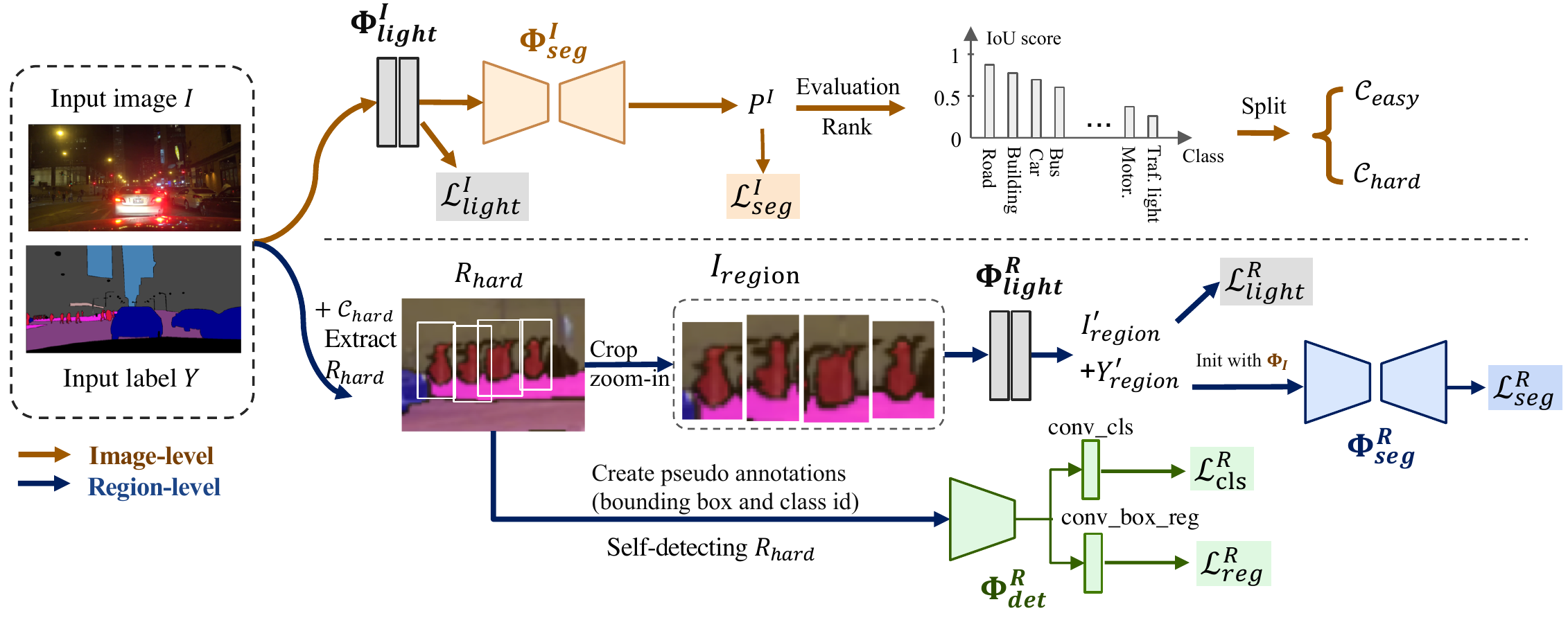}
    \vspace{-15pt}
    
   \caption{\textbf{Training pipleline}. \texttt{NightLab} training framework. The image-level modules (Upper) $\Phi_{light}^I, \Phi_{seg}^I$  will be first trained by their corresponding losses $\mathcal{L}_{light}^I, \mathcal{L}_{seg}^I$ (Sec.~\ref{sec:in_level}). After prediction over a split validation set, evaluation and ranking will be performed to split the classes into easy and hard categories  $\mathcal{C}_{easy}, \mathcal{C}_{hard}$. 
   Then, we adopt the semantic ground truth from $\mathcal{C}_{hard}$ to extract the hard regions (white boxes) by finding connected components (red regions). They can be first used to crop and zoom-in our training images to form a regional train set ($\mathcal{I}_{R}$ and$\mathcal{Y}_{R}$  ), which is utilized to optimize region-level modules (Lower) $\Phi_{light}^R, \Phi_{seg}^R$ with regional losses $\mathcal{L}_{light}^R, \mathcal{L}_{seg}^R$. Finally, based on hard regions, we derive pseudo ground truth for $\Phi_{det}^R$ (RDN or HDM ) which can be trained with regression and classification losses, i.e. $\mathcal{L}_{reg}^{R}, \mathcal{L}_{cls}^{R}$. (Sec.~\ref{sec:rdn} and Sec.~\ref{sec:serdn})}
    
    \label{fig:pipeline}
\end{figure*}

\section{NightLab}
\label{sec:method}
\vspace{-5pt}
As shown in Fig.~\ref{fig:deployment}, \texttt{NightLab} consists of two levels: image level and region level. At each level, there is a segmentation module with a network $\Phi_{seg}$. To improve the generalization of the segmentation models for night images, a ReLight Adaptation Module (ReLAM), $\Phi_{light}$, is used at each level. Region-level module works solely on hard regions in night images, which are detected by a region proposal network $\Phi_{det}^R$, \ie, region detection network (RDN) or Hardness Detection Module (HDM). Segmentation results from the two levels are merged to create the final results.

In this section, we elaborate the proposed \texttt{NightLab} architecture as follows: In section~\ref{sec:in_level} we describe the core modules at each level, \ie, ReLAM and segmentation networks. We then introduce two region proposal networks which propose hard regions to the region-level segmentation module during inference: RDN as the baseline method in section~\ref{sec:rdn}, and its improvement, HDM, in section~\ref{sec:serdn}.

\subsection{Core modules at image and region levels}
\label{sec:in_level}

\noindent\textbf{Regularized Light Adaptation Module (ReLAM).} ReLAM contains a generator to adapt the image light, and a discriminator for training. For generator we adopt the style transfer network proposed in ~\cite{Johnson2016neural_style_transfer}, which contains 6 level of resnet layers as designed~\footnote{\url{https://github.com/jcjohnson/fast-neural-style}}.  We denote the generator as  $\Phi_{light}$ which tries to align the lighting of nighttime to daytime images. It takes an RGB night image $I$ as input and output a RGB light shift $L = \Phi_{light}(I)$. The final enhanced image can be denoted as  $I^{\prime}=I+L$. 

To train ReLAM generator $\Phi_{light}$, we induce two objectives based on our collected day time image set $\mathcal{I}_d$ and night time image set $\mathcal{I}_n$. The first objective is structural  similarity loss (SSIM)~\cite{wang2004ssim, iccv2021regularizing, wu2021dannet} which penalize a dramatic change of internal texture. The second is GAN loss~\cite{goodfellow2014generative} which transfer the image for easy appearance distinguishing. 
Formally, for SSIM, we have the loss defined as, 
\vspace{-5pt}
\begin{equation}
\begin{matrix}
\mathcal{L}_{S}=\sum_{I_i \in \mathcal{I}_d, \mathcal{I}_n} 1-\text{SSIM}(I_i, I_i^{\prime}), 
\end{matrix}
\vspace{-5pt}
\end{equation}
where $I_i^{\prime}$ is the adapted image of $I_i$. 

For GAN loss,  ReLAM uses a discriminator $D$ to distinguish if the adapted image is close to daytime or nighttime following the GAN training pipeline, which can be formulated as,
\vspace{-5pt}
\begin{equation}
\begin{matrix}
      \mathcal{L}_{P}(D)=\sum\limits_{I_i \in \mathcal{I}_d, I_i^{\prime} \in \mathcal{I}^{\prime}} \log(D(I_i))+\log(1-D(I_i^{\prime}))  
\end{matrix}
\vspace{-5pt}
\end{equation}
and our final loss for ReLAM is $\mathcal{L}_{light} = \mathcal{L}_{S} + \mathcal{L}_{P}(D)$. 

ReLAM will be trained at both image ($\Phi_{light}^I$) and region level ($\Phi_{light}^R$) with their corresponding training set. In our experiments, directly using a CycleGAN~\cite{zhu2017cyclegan} image transfer could harm the segmentation performance since it generates images with lots of texture distortion, while ReLAM works better thanks to the regularization inside the architecture and losses. 

\noindent\textbf{Image-level segmentation module.} The architecture of $\Phi^I_{seg}$ is a network composed of a encoder based on Swin-Transformer~\cite{liu2021swin} and a decoder based on UperNet~\cite{eccv2018upernet}. Additionally, to increase the context modelling ability, we replace convolutional layers in UperNet with DeformConv~\cite{dai2017deformable}. We name this as ``NightLab-Baseline'', which is the best baseline we could obtain as a single network. 
To train such an architecture, we use the enhanced images and their corresponding ground truths. Formally, we adopt 2D cross entropy loss  $\mathcal{L}^{I}_{seg}$ to compare the segmentation prediction $ P^{I}$ and the ground truth annotation $Y^I$, with the same training setting as Swin-Transformer.


\noindent\textbf{Region-level segmentation module.} As we explained earlier, due to variations like lighting and scales, using only image-level segmentation module is not sufficient as some objects require different context (\eg smaller objects and low-light regions will benefit from zoomed-in views). To solve this issue, we use a region-level segmentation module that focuses on hard regions $R_{hard}$, \ie, regions at which the image-level module fails. We adopt the same architecture for region-level segmentation network $\Phi^{R}_{seg}$ as $\Phi^I_{seg}$, with the exception of the number of classes, which is dependent on the number of hard classes determined by the auto selection process described below.

We extract the hard regions $R_{hard}$ via a simple yet efficient auto selection process which utilizes the segmentation masks predicted by the image-level module $\Phi^I_{seg}$: Based on the initial segmentation prediction $P^I$ from $\Phi^{I}_{seg}$, the semantic classes $ \mathcal{C} $ are split into two sets, the easy set ${\mathcal{C}_{easy}}$ and the hard set ${\mathcal{C}_{hard}}$. Specifically, as shown in Fig.~\ref{fig:pipeline}, the per-class IoU of $P^{I}$ are first computed, and then the classes with low IoU scores ($<0.5$) are selected as ${\mathcal{C}_{hard}}$ and the rest classes consists of ${\mathcal{C}_{easy}}$. Next the instances of ${\mathcal{R}_{hard}}$ are generated based on the label masks. Since we do not have the instance-level segmentation in the ground truth annotations, we approximately consider every connected component from a class as an instance of that class. For each instance, we crop an image using a bounding box around it, which serves as the context of the instance. These cropped images are ``zoomed-in" to predefined sizes before being fed to the region-level network $\Phi^{R}_{seg}$ for training.
With the dataset, $\Phi^{R}_{seg}$ can be learnt using cross entropy loss $\mathcal{L}^{R}_{seg}= \sum_{P_j^{R} \in \mathcal{P}^{R}, Y^{R}_j\in \mathcal{Y}_{R} } CE(P_j^{R}, Y^{R}_j) $, where $P_j=\Phi_{R}(I_j^{R}), I_j^{R} \in \mathcal{I}_R$, and $\mathcal{I}_{R}$ and $\mathcal{Y}_{R}$ represent the cropped images and their semantic label masks.
\definecolor{ultramarine}{rgb}{0, 0.125, 0.476}
\begin{figure}
    \centering
    \includegraphics[width=\linewidth]{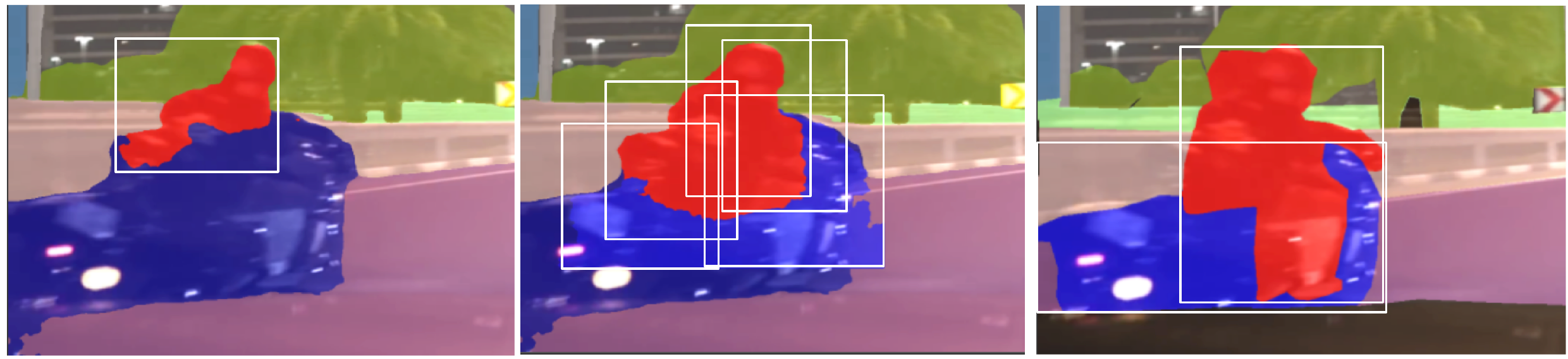}
    \vspace{-20pt}
    \caption{ Comparison of hard regions from different strategies (white boxes). Left: using prediction from $\Phi_{seg}^I$; Middle: using region detection network (RDN); Right: using ground truth. ({\color{red}{red}}: rider, {\color{blue}blue}: motorcycle, {\color{ultramarine} dark blue}: car ).}
     \vspace{-5pt}
     \label{fig:rois}
\end{figure}


        

\subsection{Self-detecting hard regions at night}
\vspace{-5pt}
\label{sec:rdn}
Region-level module cannot be directly used at inference when ground truth annotations are not available for cropping hard regions. One solution is to use prediction from $\Phi^{I}_{seg}$ to create ${\mathcal{R}_{hard}}$. The issue of this approach is that the prediction of $\Phi^{I}_{seg}$ is not always accurate. 
Instead we train a region detection network (RDN) $\Phi^R_{det}$ to detect instances in ${\mathcal{R}_{hard}}$. Inspired by Faster RCNN~\cite{ren2015faster-rcnn}, we adopt idea of the region proposal network (RPN) in ~\cite{ren2015faster-rcnn} to detect the hard regions:
we first produce the pseudo annotation boxes and label them with hard or easy based on quality of the prediction of $\Phi^{I}_{seg}$, and then RDN is learned by optimizing the objective $\mathcal{L}_{det}^{R}$, which is the sum of the bounding box regression loss and the classification loss:
\begin{align}
    \mathcal{L}_{det}^{R} &=  \mathcal{L}_{reg}^{R} + \mathcal{L}^{R}_{cls} \\
    \mathcal{L}_{reg}^{R}&= \sum\nolimits_{r_k \in \mathcal{R}_{hard}} \text{smooth\_{l1}}( t(r_{k}), t(\hat{r}_{k})) \nonumber \\
         \mathcal{L}^{R}_{cls}&= \sum\nolimits_{r_k \in \mathcal{R}_{hard}} CE(p_{r_k}, y_{r_k})
\end{align}
where $t(.)$ is a tuple that represents a bounding box, \ie, $(x,y,dw, dh)$, and $y_{r_k} $ denotes the classification label for the box derived from semantic label mask. We keep 10 proposals for each image using non maximum suppression.

As shown in Fig.~\ref{fig:rois}, our proposed RDN address the concerns. The results cover most area of rider and motorcycle even though each proposal covers only part of the object. Once $\mathcal{R}_{hard}$ is learned, images can be cropped and zoom-in, later fed through $\Phi^R_{seg}$ to produce regional estimation for $\mathcal{C}_{hard}$. At last, we can merge results from $\mathcal{P}^R$ to $\mathcal{P}^I$ to create a new mixture prediction of $\mathcal{R}_{hard}$  and $\mathcal{R}_{easy}$.

\begin{figure}[!htp]
    \centering
    \includegraphics[width=\linewidth]{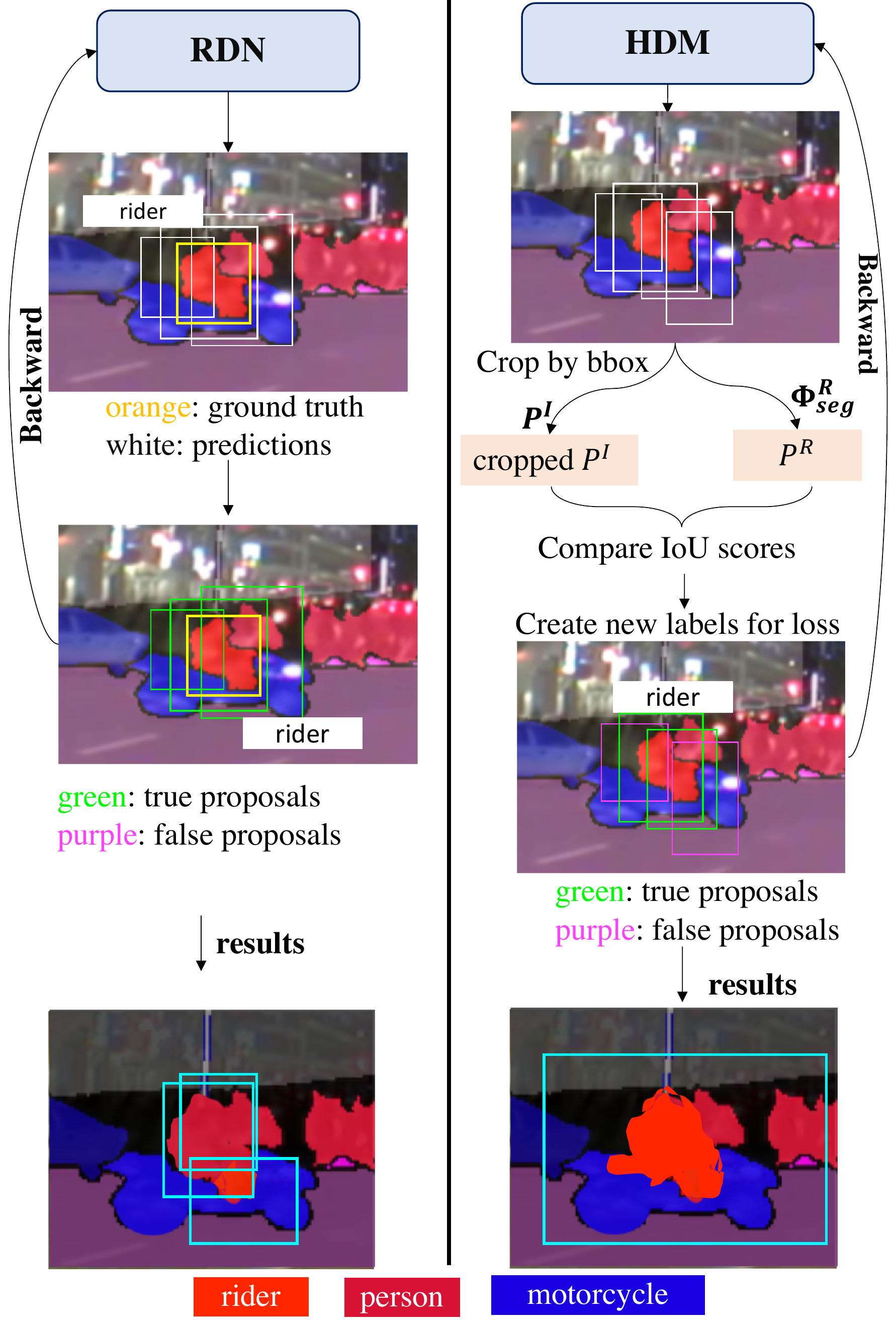}
      \vspace{-20pt}
    \caption{Region proposal networks. Left: region detection network (RDN) results of hard regions using ground truth solely based on segmentation mask. Right: hardness detection module (HDM) results using generated ground truth guided by region-level models.}
    \vspace{-10pt}
    \label{fig:SeRDN}
\end{figure}

\subsection{Detecting hard regions with contexts}
\vspace{-5pt}
\label{sec:serdn}

The problem of RDN is that RDN tends to generate a lot of proposals most of which do not contain the contexts that the region-level network $\Phi^R_{seg}$ needs. For example, as shown at the bottom left of Fig.~\ref{fig:SeRDN}, most of the proposals of ``rider'' from RDN only cover the rider but not motorcycle. Images cropped according to these proposals will be misclassified as ``person'' by $\Phi^R_{seg}$. In this section, we propose our improvement over RDN, Hardness Detection Module (HDM), which is learned to propose regions with contexts that favor the prediction of $\Phi^R_{seg}$.

Before jumping into the details of HDM, let's first explain what a ``good proposal" is: A proposal is good if $\Phi^R_{seg}$'s prediction to its region is better than the image-level model $\Phi^I_{seg}$'s prediction to the same region; in other words, a good proposal should help $\Phi^R_{seg}$ improve $\Phi^I_{seg}$'s prediction to its region. 

Follow the above intuition, we modify the learning routine of RPN in RDN as follows: for each proposal, we crop the image according to the proposal and obtain its segmentation prediction $P^R$ from $\Phi^R_{seg}$. The same proposal is used to crop the prediction from $P^I$ by $\Phi^I_{seg}$. The proposal is considered positive if the IoU score of $P^I$ is better than that of  $P^R$, and vice versa. The RPN is then trained with the new labels. As a result, HDM tends to propose regions that have better context.

As shown at the bottom right of Fig.~\ref{fig:SeRDN}, unlike RDN, HDM manages to propose a region with both the motorcycle and the person in it and $\Phi^R_{seg}$ correctly recognizes the person as ``rider'' thanks to the correct context in the region.
\begin{figure}[htbp]
    \centering
    \includegraphics[width=\linewidth]{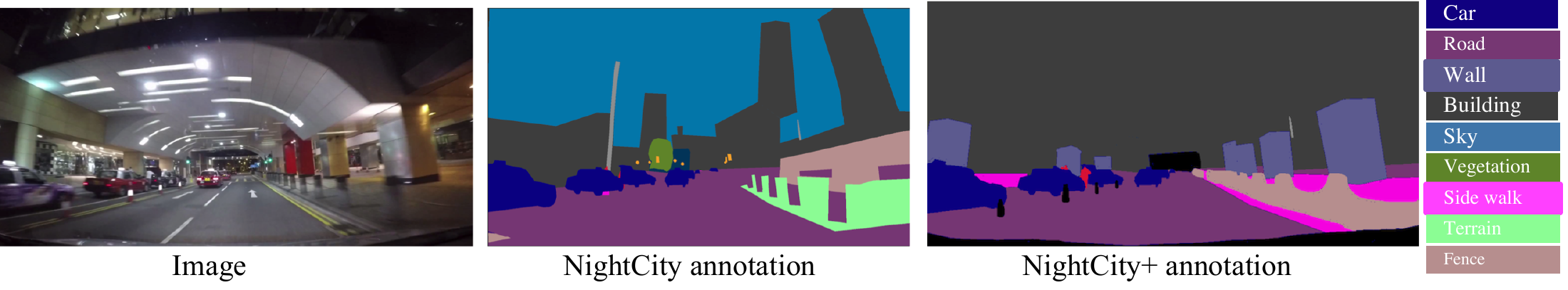}
      \vspace{-20pt}
    \caption{Example corrected label of NightCity by NightCity+.}
     \vspace{-5pt}
    \label{fig:nightcity+}
   
\end{figure}
\section{Experiments}
\vspace{-3pt}
\label{sec:experiment}

\subsection{Experimental setups}
\label{sec:setup}
\vspace{-5pt}
\paragraph{Datasets.} We consider two nighttime segmentation datasets to evaluate \texttt{NightLab}. 
First, \textbf{NightCity}~\cite{tan2020nightcity}, which is a large dataset with urban driving scenes at nighttime designed for supervised semantic segmentation. It consists of 2998/ 1299 train/val images with full pixel-wise annotations. The labels are compatible with Cityscapes~\cite{Cordts2016cityscapes} where there are 19 classes of interests in total. From the dataset, we found there are some mis-labelled validation images (Fig.~\ref{fig:nightcity+}), especially for some slim objects, which is difficult to reveal the true improvements. Therefore, we asked human labellers to relabel part of the ``validation'' set for more accurate evaluation. We call this \textbf{NightCity+},  and report all our experimental results on the new val set. More labelled details can be found in supplementary materials.  Similar with NightCity~\cite{tan2020nightcity}, we also experimented with Cityscapes as assist training data to help improve the performance.
Second, \textbf{BDD100K} ~\cite{yu2018bdd100k}, which is a high-resolution autonomous driving dataset with 100,000 video clips in multiple cities and under various conditions. 

\begin{table}[!htbp]\centering
\resizebox{\linewidth}{!}{%
\begin{tabular}{l|c|c|c|c}
\toprule
\multicolumn{5}{c}{\textbf{(a) NightCity+}}\\
Network &Backbone &Resolution & NightCity+  & w/Citys \\
\hline
\hline
*NightCity\cite{tan2020nightcity} &Res101&512x1024 & 51.5 & 53.9\\
PSPNet\cite{zhao2017pspnet} &Res101&512x1024 &54.75 & 56.89\\
PSPNet\cite{zhao2017pspnet} &Res101&1024x2048 &55.64 &57.52\\
DeeplabV3+\cite{chen2018deeplabv3plus} &Res101&512x1024 & 54.21 &58.29 \\
DeeplabV3+\cite{chen2018deeplabv3plus} &Res101&1024x2048 &54.47 &59.03 \\
UPerNet\cite{liu2021swin} &Swin-Base&512x1024 &57.71 &59.35 \\

HRNetV2\cite{wang2020hrnet} &HRNet-W48&1024x2048 & 55.89 & 58.49\\
DANet\cite{fu2019danet} &Res101&1024x2048 & 55.98 & 57.72\\

UPer-Swin\cite{eccv2018upernet}&Res101&1024x2048 & 55.81& 56.98\\
UPer-ViT \cite{dosovitskiy2020vit} & ViT&  1024x2048&57.13 &58.07\\

UPer-Swin\cite{liu2021swin} &Swin-Base&1024x2048 &58.25 &59.67 \\
\hline
*\texttt{NightLab-HDM} &Swin-Base &512x1024 & 59.84 & 61.07 \\
\texttt{NightLab} (DeeplabV3+) &Res101&1024x2048 & 56.21 & 60.41\\
\texttt{NightLab-Baseline} &Swin-Base &1024x2048 &59.25 & 60.37\\
\texttt{NightLab-RDN} &Swin-Base &1024x2048 &60.27 & 62.11\\
\texttt{NightLab-HDM} &Swin-Base &1024x2048 & \textbf{60.73} & \textbf{62.82} \\

\toprule
\multicolumn{5}{c}{\textbf{(b) BDD100K-Night}}\\
Network &Backbone & Resolution &Night & w/100K\\
\hline
\hline
PSPNet\cite{zhao2017pspnet} &Res101 &720x1280&29.96 & 46.24\\

HRNetV2\cite{wang2020hrnet} &HRNet-W48&720x1280 & 29.86&44.32 \\
DANet\cite{fu2019danet} &Res101&720x1280 & 29.46 & 42.64\\
DeeplabV3+\cite{chen2018deeplabv3plus} &Res101&720x1280 & 30.11&43.44 \\
UPerNet\cite{eccv2018upernet}&Res101& 720x1280& 30.88& 47.68\\
UPer-ViT \cite{dosovitskiy2020vit} &ViT& 720x1280 & 30.74&47.81\\
UPer-Swin\cite{liu2021swin} &Swin-Base& 720x1280&31.74 &48.04 \\

\hline
\texttt{NightLab} (DeeplabV3+) &Res101& 720x1280& 31.27 & 45.11\\
\texttt{NightLab-Baseline} &Swin-Base&720x1280 & 32.37 &48.52\\
\texttt{NightLab-RDN} &Swin-Base &720x1280 &34.13  &49.81\\
\texttt{NightLab-HDM}& Swin-Base & 720x1280 & \textbf{35.41} & \textbf{50.42}\\
\end{tabular}}
\vspace{-5pt}
\caption{Comparisons to SoTA semantic segmentation networks on NightCity+ and BDD-Night with metric of mIoU. The results of first column after ``Resolution'' are models train with only night images, and the results of the column after trained with daytime data augmentation, i.e. with Cityscapes to NightCity, and BDD100K day images to BDD100K-Night. Here, for NightCity, lines with $*$ denotes evaluation is done over the original NightCity val set since we do not have models in NightCity~\cite{tan2020nightcity}.}
\vspace{-5pt}
\label{tab:arch}
\end{table}
\begin{table*}[!htbp]\centering
\resizebox{0.8\linewidth}{!}{%
\begin{tabular}{l|c|c|c|c|c|c}
Method& Adaptation Approach&Network &Nightcity+ &Nightcity+ and Citys&BDD100K-Night & BDD100K \\
\hline
\hline
NightCity\cite{tan2020nightcity} & Exposure-Aware& Res101 &51.8 &53.9& -& -\\
UPerNet \cite{liu2021swin}& Segmentation& UPerNet-Swin &57.71 & 59.35&31.74 &48.52\\

Pix2PixHD\cite{isola2017pix2pix} &Image Translation& UPerNet-Swin &- & 43.38& -&38.67\\
CycleGAN\cite{zhu2017cyclegan} &Image Translation  &UPerNet-Swin &- & 44.07& -& 39.64\\
SingleHDR\cite{liu2020singlehdr} &Image Enhancement  &UPerNet-Swin &57.07 & 58.88 &31.64 & 48.32\\
DANNet\cite{wu2021dannet} & Network Adaptation &UPerNet-Swin & -&58.69 & - & 48.25 \\
AdaptSeg\cite{Tsai_adaptseg_2018} & Network Adaptation  &UPerNet-Swin & -& 58.29& - & 48.32 \\
\hline
\texttt{NightLab-B} & Segmentation  &UPerNet-Swin-DeformConv &59.25 &60.37 & 32.37& 48.52\\

\texttt{NightLab-RDN}  & Dual-level segmentation  &UPerNet-Swin-DeformConv &60.27 & 62.11 & 34.13& 49.81\\
\texttt{NightLab-HDM} & Dual-level segmentation 
&UPerNet-Swin-DeformConv &\textbf{60.73} & \textbf{62.82} & \textbf{35.41}& \textbf{50.24}\\
\end{tabular}}
\vspace{-5pt}
\caption{Comparison study of adaptation approaches. mIoU(\%) are reported.}
\vspace{-5pt}
\label{tab:all_approach }
\end{table*}
\begin{figure*}[!htbp]
\includegraphics[width=\linewidth]{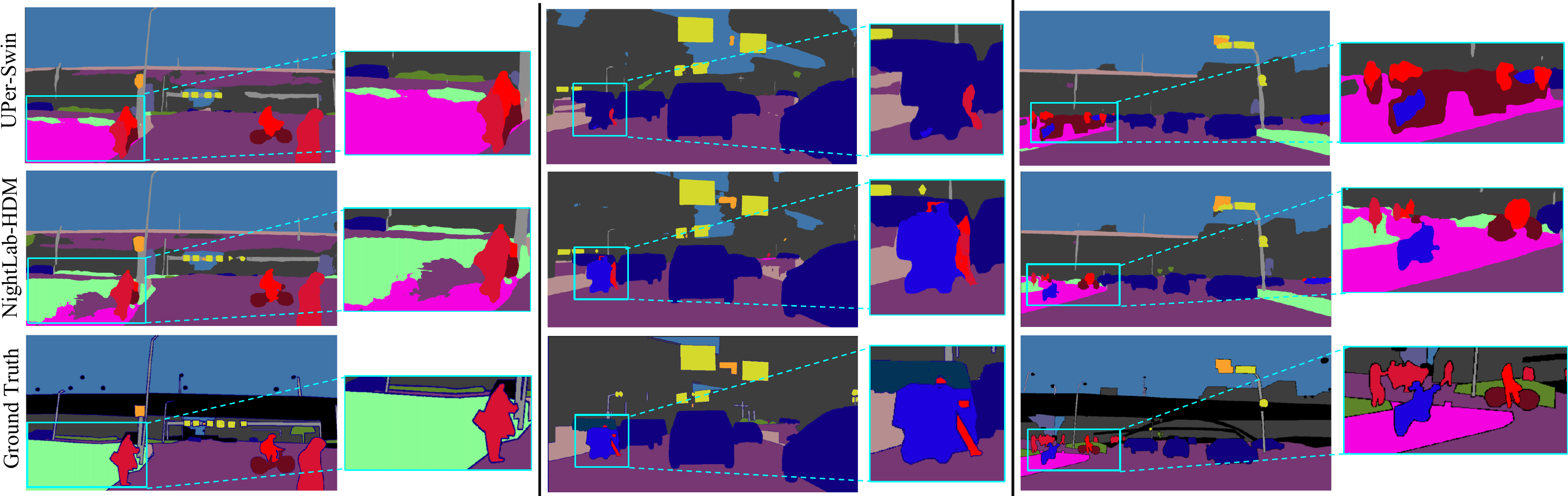}
\vspace{-20pt}
\caption{Qualitative results on NightCity+ val set jointly trained with Cityscapes. }
\vspace{-10pt}
\label{fig:qualitative}
\end{figure*}
We pick the night images with their label inside to build a new dataset, namely \textbf{BDD100K-Night}, which consists of 343/58 images in train/val sets with  18 classes of interests. The amount of data is much less than NightCity+,  to augment the training, we adopt the whole BDD100K dataset including 7,000 images and the corresponding annotations to jointly train, and then evaluate on BDD100K-Night val set. 
Last, we also explore many other datasets for setting benchmarks, such as Zurich-Dark~\cite{iccv19gcma}, ApolloScape~\cite{huang2018apolloscape}, but found them containing few or no night training images with labels, which is not suitable in our situation.

\noindent\textbf{Implementation Details.}  Since both datasets do not include official test sets, we treat their val set as test, and randomly split the original train set to train and val set with a portion of 3:1 for hyperparameter tuning, hard class selection and model selection. After the tuning, the full train set is used to optimize the final model, and test over the test set. For training and inference, the images of NightCity+ will be rescaled to $1024\times2048$.  For both datasets, we adopt training augmentations of random scale with ratio sampled in range of $(0.5, 2.0)$,  random flip,  photonmetric distortion and normalization.  Afterwards, the image will be cropped into a shape of $512\times1024$ before feeding into the model. For evaluation, we apply a multi-scale augmentation strategy with ratios of [0.25, 0.5, 0.75, 1.0, 1.25].  During training, we select hard classes with mIoU less than 0.5. 
We use CityScapes and BDD100K day split for day image set,  NightCity+ and BDD100K night split for night image set to train image-level ReLAM. For region-level ReLAM, we crop out corresponding hard classes in day and night images to compose the train set.

We run our experiments based on mmsegmentation~\cite{mmseg2020}. Our experiments are performed on 8 V100 GPUs, with 2 samples per GPU. Sync BatchNorm is turned on for all experiments. We produce the baseline results for NightCity+ and BDD100K in Tab.~\ref{tab:arch} with default hyperparameters in mmsegmentation where the models are trained with 80k iterations. \texttt{NightLab} follows the same training configurations of UPer-Swin in ~\cite{liu2021swin} for training segmentation modules. While for ReLAM, we adopt the configurations as in ~\cite{Tsai_adaptseg_2018}, and for HDM, we follow the setting in FastR-CNN~~\cite{ren2015faster-rcnn}. Since the two level model can be run in parallel, our approach runs in the same speed as Swin-Transformer.

\subsection{Experimental results}
\noindent\textbf{Compare to SoTA methods.} In Tab.~\ref{tab:arch}, we compare \texttt{NightLab} to SoTA semantic segmentation methods. For each baseline, we 
create the experimental result with the same training configurations as ours, and obtain the result by training with night data only or training with additional day data as discussed in experimental setup in Sec.~\ref{sec:setup}. 

From the table, we can see our constructed baseline, ``\texttt{NightLab-Baseline}'', simplified as \texttt{NightLab-B} latter, contains only image-level segmentation module already outperforms the best of concurrent SoTA networks, \ie, ``Uper-Swin'' based on Swin-transformer~\cite{liu2021swin}, on both datasets with gains of $\sim$1\% for both training configurations.
Specifically, for NightCity+,  ``NightLab-RDN'' represents adding RDN module inside the network, and it outperforms UPer-Swin with a margin of 2.44\% when jointly trained with Cityscapes. After replacing RDN with HDM, ``NightLab-HDM'' is able to perform better, which achieved mIoU scores of 60.73\% and 62.82\% in single train and joint train, yielding 2.48\% and 3.15\% improvements over UPer-Swin. Similar gain is observed for BDD100K-Night. ``NightLab-HDM'' achieves the best performances with 35.41\% and 50.42\% under single train and joint train settings. To further verify the effectiveness of our proposed modules, we switch the base network of NightLab-HDM to Deeplab V3+, as shown in lines of ``NightLab (DeeplabV3+)'', we observed sufficient gain over baseline ``DeeplabV3+'' with improvements of 1.74\% and 1.38\% in NightCity+, and 1.16\% and 1.67\% in BDD100K-Night. 

\begin{table*}[!htp]

  \resizebox{\linewidth}{!}{%
    \begin{tabular}{l|rrrrrrrrrrrrrrrrrrr|r}
   Method & \multicolumn{1}{l}{road} & \multicolumn{1}{l}{side.} & \multicolumn{1}{l}{build.} & \multicolumn{1}{l}{wall} & \multicolumn{1}{l}{fence} & \multicolumn{1}{l}{pole} & \multicolumn{1}{l}{light} & \multicolumn{1}{l}{sign} & \multicolumn{1}{l}{vege} & \multicolumn{1}{l}{terr.} & \multicolumn{1}{l}{sky} & \multicolumn{1}{l}{pers.} & \multicolumn{1}{l}{rider} & \multicolumn{1}{l}{car} & \multicolumn{1}{l}{truck} & \multicolumn{1}{l}{bus} & \multicolumn{1}{l}{train} & \multicolumn{1}{l}{moto.} & \multicolumn{1}{l}{bicy.} & \multicolumn{1}{l}{mIoU} \\
    \hline
    \hline
   UPerNet-Swin & 92.1  & \textbf{55.3}  & 84.4  & 59.1  & 56.1  & 38.9  & 34.0  & 60.9  & 63.1 & \textbf{29.9}  & 89.0  & 60.9  & 32.7  & 85.7  & 66.5  & 73.5  & 60.1  & 39.2  & 45.7  & 59.35 \\
  SingleHDR & 90.8  & 51.7 & 83.1 & 59.0 & 53.4 & 34.9 & 34.2 & 57.1 & 60.4 & 27.5 & 86.5 & 55.4 & 34.0 & 80.9 & 66.5 & 73.2 & 57.9 & 39.0 & 38.7 & 57.07 \\
Pix2PixHD & 85.9 & 33.5 & 68.8 & 50.0 & 42.9 & 27.0 & 13.8 & 34.6  & 47.9& 20.1 & 82.8 & 35.5 & 12.1 & 72.7 & 53.5 & 58.3 & 42.6 & 17.3 & 25.0  & 43.38 \\

 DANNet &91.5  & 53.8  & 85.4  & 59.9  & 54.9  & 38.9  & 34.7  & 60.0  & 62.2  & 28.7  & 88.2  & 58.1  & 35.4  & 83.1  & 66.9  & 72.1  & 58.1  & 40.6  & 42.0  & 58.69 \\ 
    \hline

\texttt{NightLab-B} &92.4  & 54.2  & 85.3  & 59.3  & 57.3  & 38.2  & 28.3  & 61.8  & 62.3  & 24.0  & 89.1  & 62.4  & 43.2  & 86.0  & 68.1  & 78.9  & 61.4  & 48.6  & 46.2  & 60.37 \\
 \texttt{NightLab-RDN}&   92.5  & 53.6  & 85.2  & 59.9  & 58.0  & 42.2  & 37.7  & 62.9  & \textbf{63.4}  & 26.6  & \textbf{89.7}  & 63.3  & 45.2  & 86.4  & 70.1  & 80.5  & 62.6  & 50.2  & 50.6  & 62.11 \\
  \texttt{NightLab-HDM}& \textbf{ 92.6}  & 54.9  & \textbf{85.8}  & 59.1  & \textbf{58.4}  & \textbf{43.1}  & \textbf{38.1}  & \textbf{63.3}  & 63.0  & 26.6  & 89.3  & 63.3  & \textbf{47.1}  & \textbf{86.7}  & \textbf{71.9}  & \textbf{81.0}  & \textbf{63.7}  & \textbf{52.3}  & \textbf{54.8}  &  \textbf{62.82} \\

    \end{tabular}}
    \vspace{-10pt}
      \caption{Per class iou scores. Model are jointly trained with NightCity+ and Cityscapes, evaluated on NightCity+ val set.   }
      \vspace{-10pt}
  \label{tab:class_iou}%
\end{table*}%

\noindent\textbf{Compare to adaptation for segmentation methods.} Since most existing methods for segmentation through adaption are unsupervised, directly comparing with them on our benchmarks is not fair. Therefore,  we consider to train an adaptor to adapt night images to day images, then use the adapted night dataset plus real day dataset to supervise a segmentation model based on UPerNet-Swin. We hope the adaptation can help the segmentation network learn better.
In Tab.~\ref{tab:all_approach }, we explore various SoTA adaptors to adapt night images to day images for training. 
Specifically, we first use ``Pix2PixHD/CycleGAN'' to transfer the appearance of the nighttime images into daytime. 
However, under supervised setting, we found such adaptation actually performs worse than a vanilla baseline. This is because the adapted appearance of night images are dramatically changed, which can hardly be consistent in training and testing time.
Then, we additionally explore other DA methods with less modification on image contents by using a pretrained SingleHDR~\cite{liu2020singlehdr} for image enhancement, although it does not harm the results much, we do not see any improvements. 

Finally, we explore whether unsupervised adapted networks can provide better pre-trained feature for night images since the weight itself should contain adaptation ability. Specifically, we adopt DANNet~\cite{wu2021dannet} and AdaptSeg~\cite{Tsai_adaptseg_2018} to first train an adapted segmentation network with unlabelled day/night images, and then finetune it with our full labelled day/night dataset. Unfortunately, it also does not help with the accuracy as shown in lines of ``DANNet'' and ``AdaptSeg''. It seems the adapted feature could be biased, yielding a slightly worse optimized weights than vanilla training in our experiments. 
More details can be found in supplementary materials. 

\noindent\textbf{Class performance} 
We further analyze the model contributions for each class shown in Tab.~\ref{tab:class_iou}. We can see \texttt{NightLab} makes improvements on almost all classes. Especially, hard classes detected such as pole, rider, motorcycle, and bicycle are improved mostly thanks to HDM,  ReLAM and region-level modules. For example, the score of class \textit{pole} is increased from 38.2\% to 42.2\% after applied HDM and region-level model, which are within our expectation. Corresponding qualitative results are shown in Fig.~\ref{fig:qualitative}.  
However, there are some hard classes detected have not been improved such as \textit{terrain}. This is due to the fact that terrain is more likely to be background rather than object, which can be easily confused by \textit{vegetation} in particular in the dark environment. We found the approach is more effective for object-like classes. 
  
\begin{table}[!htp]\centering

\resizebox{\linewidth}{!}{%
\begin{tabular}{l|c|l|l|c|c}

Method &Backbone & FPN &Seg head & Nightcity+ &+Citys \\
\hline
\hline
UPerNet\cite{eccv2018upernet} &Res101 &Conv2D  & Conv2D & 55.81& 59.03\\
\texttt{NightLab-B} &Res101 & Conv2D & DefConv\cite{dai2017deformable} &56.31  & 59.33 \\
\texttt{NightLab-B} & Res101 & DefConv\cite{dai2017deformable} &  DefConv\cite{dai2017deformable} & 56.54 & 59.85\\
\hline
 UPer-Swin\cite{liu2021swin} &Swin-Base &Conv2D &  Conv2D&58.25 &59.67 \\
\texttt{NightLab-B} &Swin-Base & Conv2D &  DefConv\cite{dai2017deformable}&58.68 &59.99 \\
\texttt{NightLab-B} &Swin-Base & DefConv\cite{dai2017deformable} & DefConv\cite{dai2017deformable} & \textbf{59.25} & \textbf{60.37}\\
\end{tabular}}
\vspace{-10pt}
\caption{Ablation study on our proposed baseline architectures adding deformable convolution (DefConv~\cite{dai2017deformable}) to enrich contextual features for multiscale objects. Results are reported on NightCity+ val set.}\label{tab:ablation_arch }
\vspace{-10pt}
\end{table}
\subsubsection{Ablation study}

\vspace{-5pt}
\noindent\textbf{\texttt{NightLab-B} architecture} We present the ablation study of the baseline architecture shown in Tab.~\ref{tab:ablation_arch }. We make modification of Conv2D of the decode head (UPerHead from UPerNet~\cite{eccv2018upernet}) to build our baseline method. UPerHead is composed of a feature fusion module FPN and a segmentation conv head. Deformable conv can be a substitute of regular conv to produce feature with better context. The proposed decode head can be combined with any backbone. We verify the effectiveness by performing experiments on model with a backbone of Swin-Transformer and ResNet101. As shown in the table, Deformable conv results in better performances with both backbones. When replacing all the conv layers with deformable conv, the model achieves the best performance.

\begin{table}[!htp]

\resizebox{1.0\linewidth}{!}{%
\begin{tabular}{l|c}
Method & mIoU(\%)\\
\hline
\hline
UPerNet-Swin[\cite{liu2021swin}] & 59.67 \\
\texttt{NightLab-B}  &60.37 \\

\hline
\texttt{NightLab-B} + $\Phi^{R}_{seg}$ w/ RDN proposed hard regions  &61.51  \\
\texttt{NightLab-B} + $\Phi^{R}_{seg}$ w/ HDM proposed hard regions  &62.31   \\
\hline
\texttt{NightLab-B} + $\Phi_{light}^I$ &60.74 \\
\texttt{NightLab-B} + $\Phi_{light}^I$ + $\Phi^{R}_{seg}$ w/ RDN proposed hard regions  &61.87  \\
\texttt{NightLab-B} + $\Phi_{light}^I$ + $\Phi^{R}_{seg}$ w/ HDM proposed hard regions& 62.51  \\
\hline
\texttt{NightLab-B} + $\Phi_{light}^I$ + $\Phi^{R}_{seg}$ w/ RDN proposed hard regions + $\Phi_{light}^R$ &62.11  \\
\texttt{NightLab-B} + $\Phi_{light}^I$ + $\Phi^{R}_{seg}$ w/ HDM proposed hard regions + $\Phi_{light}^R$ &62.82  \\

\end{tabular}%
}
\vspace{-5pt}
\caption{Ablation study on \texttt{NightLab} model variants. Models are trained jointly with NightCity+ and Cityscapes, evaluated on NightCity+ val set. \texttt{NightLab-B} represents our proposed baseline segmentation architecture. }\label{tab:ablation_rpn }
\vspace{-5pt}
\end{table}

\noindent\textbf{\texttt{NightLab} modules} We present the ablation study for each module of \texttt{NightLab} in Tab.~\ref{tab:ablation_rpn }. We first show the effectiveness of the dual-level architecture without lighting adaptation. Utilizing $\Phi_{seg}^{R}$ with hard regions detected by RDN or HDM results can raise the performance from 60.37\% to 61.51\% and 62.01\% respectively. HDM provides the best result. Next, we demonstrate the lighting adaptation module of ReLAM. We can see both levels of ReLAM raise the performance for all modules.  We observe that adding the lighting adaptation module towards the whole image can increase the mIoU by $\sim$0.5\%. A further region based lighting adaptation module can raise slight improvement of $\sim$0.3\%.


\section{Conclusion}
This paper presents \texttt{NightLab},  an architecture that is suitable for night scene segmentation. It contains dual-level models, which segments images with proper context and lights in a supervised setting, yielding SoTA performance. However, the overall performance of segmentation accuracy at night is still far behind that from daytime. This work takes a few steps in effectively mining context and reduce light variations in challenging visual situation, and we hope it may motivate other researchers to discover other crucial properties toward closing the performance gap at night.
{\small
\bibliographystyle{ieee_fullname}
\bibliography{egbib}
}

\end{document}